\documentclass[11pt,a4paper]{article}
\usepackage[hyperref]{emnlp2020}
\usepackage{times}
\usepackage{latexsym}
\usepackage{booktabs}
\usepackage{graphicx}
\usepackage{amsmath}
\usepackage{amsfonts}

\usepackage{microtype}

\aclfinalcopy 
\def\aclpaperid{1566} 

\title{Denoising Relation Extraction from Document-level Distant Supervision}

\author{Chaojun Xiao\textsuperscript{\rm 1},
Yuan Yao\textsuperscript{\rm 1},
Ruobing Xie\textsuperscript{\rm 2},
Xu Han\textsuperscript{\rm 1},
Zhiyuan Liu\textsuperscript{\rm 1}\thanks{~~Corresponding author.} \\
\textbf{Maosong Sun\textsuperscript{\rm 1}, 
Fen Lin\textsuperscript{\rm 2},
Leyu Lin\textsuperscript{\rm 2}}\\
\textsuperscript{\rm 1}Department of Computer Science and Technology\\
Institute for Artificial Intelligence, Tsinghua University, Beijing, China\\
Beijing National Research Center for Information Science and Technology, China \\
\textsuperscript{\rm 2}WeChat Search Application Department, Tencent, China\\
{\small\tt \{xcjthu,thu.hanxu13\}@gmail.com, \{yaoyuanthu,xrbsnowing\}@163.com} \\
{\small\tt \{lzy,sms\}@tsinghua.edu.cn, \{felicialin,goshawklin\}@tencent.com}
\\}

\date{}

\begin{document}
\maketitle
\begin{abstract}

Distant supervision (DS) has been widely used to generate auto-labeled data for sentence-level relation extraction (RE), which improves RE performance. However, the existing success of DS cannot be directly transferred to the more challenging document-level relation extraction (DocRE), since the inherent noise in DS may be even multiplied in document level and significantly harm the performance of RE. To address this challenge, we propose a novel pre-trained model for DocRE, which denoises the document-level DS data via multiple pre-training tasks. Experimental results on the large-scale DocRE benchmark show that our model can capture useful information from noisy DS data and achieve promising results. The source code of this paper can be found in \url{https://github.com/thunlp/DSDocRE}.
\end{abstract}


\section{Introduction}

Relation extraction (RE) aims to identify relational facts between entities from texts. Recently, neural relation extraction (NRE) models have been verified in sentence-level RE \cite{zeng2014relation}. Distant supervision (DS) \cite{mintz2009distant} provides large-scale distantly-supervised data that multiplies instances and enables sufficient model training.

\begin{figure}
    \centering
    \includegraphics[width=\linewidth]{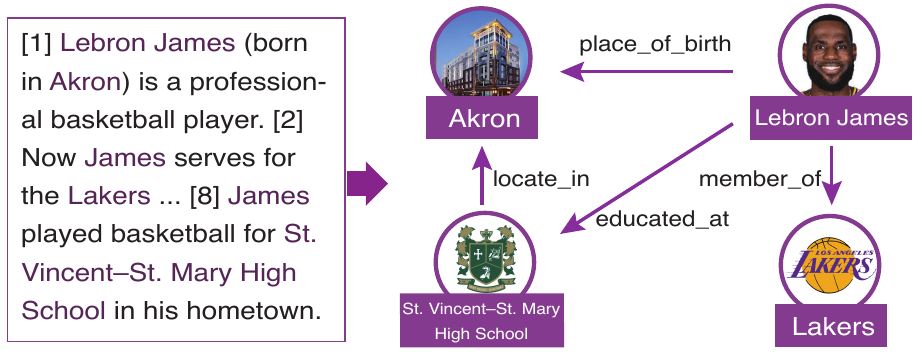}
    \caption{An example of DocRE. Given a document, DocRE models should capture the relational semantics across sentences to extract multiple relational facts.}
    \label{fig:example}
\end{figure}

Sentence-level RE focuses on extracting \emph{intra-sentence} relations between entities in a sentence. However, it is extremely restricted with generality and coverage in practice, since there are plenty of \emph{inter-sentence} relational facts hidden across multiple sentences. Statistics on a large-scale RE dataset constructed from Wikipedia documents show that at least $40.7\%$ relational facts can only be inferred from multiple sentences \cite{yao2019docred}.
Therefore, \emph{document-level relation extraction (DocRE)} is proposed to jointly extract both inter- and intra- sentence relations \cite{christopoulou2019connecting}.
Fig.~\ref{fig:example} gives a brief illustration of DocRE.

Most DocRE models heavily rely on high-quality human-annotated training data, which is time-consuming and labor-intensive. However, 
it is extremely challenging to extend the sentence-level DS to the document level. The challenges of conducting document-level DS mainly come from: (1) Each entity contains multiple mentions, and mentions without relational context bring noise to entity representations; (2) The inherent noise of DS will be even multiplied at the document level. Statistics in \citet{yao2019docred} show that $61.8\%$ inter-sentence relation instances generated by document-level DS are actually noise; (3) It is challenging to capture useful relational semantics from long documents, since most contents in the documents may be irrelevant to the given entities and relations. 
In sentence-level RE, several efforts~\cite{lin2016neural,DBLP:conf/aaai/FengHZYZ18} have been devoted to denoise the DS corpus by jointly considering multiple instances. However, these denoising methods can not be directly adapted to DocRE, since they are specially designed for bag-level RE evaluations.

In this work, we attempt to introduce document-level DS to DocRE after denoising.
To alleviate the noise, we propose a pre-trained model with three specially designed tasks to denoise the document-level DS corpus and leverage useful information. The three pre-training tasks include:
(1) \textbf{Mention-Entity Matching}, which aims to capture useful information from multiple mentions to produce informative representations for entities. It consists of intra-document and inter-document sub-tasks. The intra-document sub-task aims to match masked mentions and entities within a document to grasp the coreference information. The inter-document sub-task aims to match entities between two documents to grasp the entity association across documents.
(2) \textbf{Relation Detection}, which focuses on denoising ``Not-A-Relation (NA)" and incorrectly labeled instances by detecting the entity pairs with relations, i.e., positive instances. It is specially designed as the document-level denoising task. We also conduct a pre-denoising module trained with this task to filter out NA instances before pre-training.
(3) \textbf{Relational Fact Alignment}, which requires the model to produce similar representations for the same entity pair from diverse expressions. This allows the model to focus more on diverse relational expressions and denoising irrelevant information from the document.



In experiments, we evaluate our model on an open DocRE benchmark and achieve significant improvement over competitive baselines. We also conduct detailed analysis and ablation test, which further highlight the significance of DS data and verify the effectiveness of our pre-trained model for DocRE. To the best of our knowledge, we are the first to denoise document-level DS with pre-trained models. We will release our codes in the future.


\section{Related Work}

\textbf{Sentence-level RE.} Conventional NRE models focus on sentence-level supervised RE~\cite{zeng2014relation,takanobu2019hierarchical}, which have achieved superior results on various benchmarks.
Other approaches focus on using more data with distant supervision mechanism~\cite{mintz2009distant,min2013distant}. To denoise distantly supervised corpus, they introduce attention~\cite{lin2016neural,zhou2018distant}, generative adversarial training~\cite{qin2018dsgan} and reinforcement learning~\cite{DBLP:conf/aaai/FengHZYZ18} to select informative instances. It is hard to directly adopt these models to DocRE, since DocRE should extract multiple relational facts from each document. \citet{soares2019matching} propose a pre-trained model for sentence-level RE.

\smallskip
\noindent
\textbf{Document-level RE.} Document-level RE attempts to extend the scope of knowledge acquisition to the document level, which has attracted great attention recently~\cite{yao2019docred}. Some works use linguistic features~\cite{xu2016cd,gu2017chemical} and graph-based models~\cite{christopoulou2019connecting,sahu2019inter} to extract inter-sentence relations on human-annotated data. \citet{quirk2017distant} and \citet{peng2017cross} attempt to extract inter-sentence relations with distantly supervised data. However, they only use entity pairs within three consecutive sentences.
Different from these works, we bring in document-level DS to DocRE and conduct pre-training to denoise these DS data.

\section{Methodology}

\begin{figure}
    \centering
    \includegraphics[width=\linewidth]{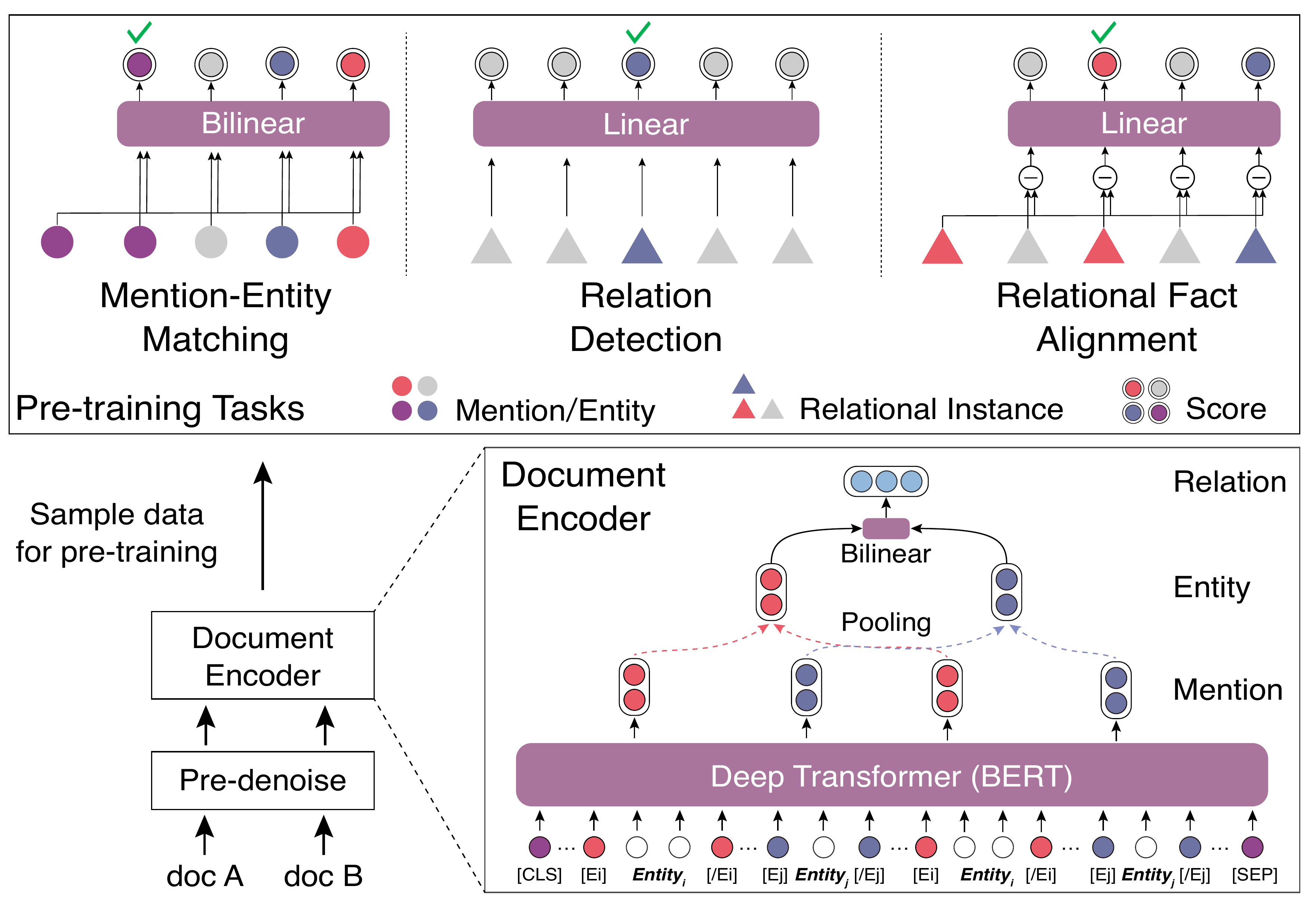}
    \caption{The framework of our proposed model.}
    \label{fig:encoder}
\end{figure}

In this section, we present our proposed model in detail.
Fig.~\ref{fig:encoder} gives an illustration of our framework.
We first apply the pre-denoising module to screen out some NA instances from all documents. Then we pre-train the document encoder with three pre-training tasks on the document-level distantly supervised dataset. Finally, we fine-tune the model on the human-annotated dataset.

\subsection{Document Encoder}
\label{sec:encoder}

We adopt BERT \cite{devlin2019bert} as the document encoder to encode documents into representations of entity mentions, entities and relational instances.
Let $D = \{\omega_i\}_{i=1}^{n}$ denote the input document which consists of $n$ tokens, and $V = \{e_i\}_{i=1}^{|V|}$ be the set of entities mentioned in the document, where entity $e_i = \{m_{i}^j\}_{j=1}^{l_i}$ contains $l_i$ mentions in the document.
Following \citet{soares2019matching}, 
we use entity markers \texttt{[Ei]} and \texttt{[/Ei]} for each entity $e_i$. The start marker \texttt{[Ei]} is inserted at the begin of all mentions of entity $e_i$, and the end marker \texttt{[/Ei]} is inserted at the end. 

We use BERT to encode the token sequence with entity markers into a sequence of hidden state $\{\mathbf{h}_1,...,\mathbf{h}_{\hat{n}}\}$, where $\hat{n}$ indicates the length of the sequence with entity markers. We define representation $\mathbf{m}_i^j$ of each entity mention as the hidden state of its start marker.
Then a max-pooling operation is performed to obtain the aggregated representation of entity $e_i$ from its mentions: $\mathbf{e}_i = \text{MaxPooling}(\{\mathbf{m}_{i}^{j}\}_{j=1}^{l_i})$.
Next, for each entity pair $(e_i, e_k)$, we use a bilinear layer to compute the relational representation: $\mathbf{r}_{i,k} = \text{Bilinear}_{E}(\mathbf{e}_i, \mathbf{e}_k)$.

\subsection{Pre-training Tasks}

We design three pre-training tasks, which help the model to denoise document-level DS data and learn informative representations in both mention/entity-level and relation-level from large-scale DS data.

\smallskip
\noindent
\textbf{Mention-Entity Matching.} An entity is usually mentioned multiple times in a document, and it is important for expressive entity representations to capture relational information from these mentions.
Hence, we propose the mention-entity matching task to help the model to produce expressive representations for mentions and entities, which includes intra-document and inter-document sub-tasks.

The intra-document sub-task requires the model to 
grasp the coreference information within a document. We randomly mask an entity mention and require the model to predict which entity in the document it belongs to. Formally, given the masked entity mention $m^{q}$ and $k_m$ entities $\{e_m^i\}_{i=1}^{k_m}$ from the same document, we compute the matching score for $e_m^i$ and $m^q$ with a bilinear layer as follows:
\begin{equation}
    s_m(e_m^i, m^q) = \text{Bilinear}_M(\mathbf{e}_m^i, \mathbf{m}^q).
\end{equation}

The inter-document sub-task requires the model to link the same entity in two different documents. It aims to develop the model to encode useful information from the contexts into the representations. 
Given the entities $\{e_A^i\}_{i=1}^{k_e}$ from document $d_A$ where $k_e$ is the size of the entity set, and the entity $e^q_B$ from document $d_B$ which is also mentioned in $d_A$, we define the matching score for entity $e_B^q$ and $e_A^i$ as:
\begin{equation}
    s_m(e_A^i, e_B^q) = \text{Bilinear}_M(\mathbf{e}_A^i, \mathbf{e}_B^q),
\end{equation}
where $\text{Bilinear}_M$ indicates the same bilinear layer in intra-document sub-task. 
Then both matching scores are fed into an output softmax function.


\smallskip
\noindent
\textbf{Relation Detection.} 
The NA relation is dominating in DocRE. It is necessary for models to denoise NA instances and to identify the true positive instances from NA noise.
Therefore, we design this task, which requires the model to distinguish positive entity pairs from NA instances.
Formally, given $k_n$ instances $\{r_n^i\}_{i=1}^{k_n}$ from given documents where only one is positive, we have their positive score as:
\begin{equation}
    s_n(r_n^i) = \mathbf{w}_n\mathbf{r}_n^i + b_n,
    \label{score:positive}
\end{equation}
where $\mathbf{w}_n$ and $b_n$ indicate weights and bias. Next, we apply a softmax function to compute the probability of $i$-th instance to be positive.

Similar to the previous mention/entity-level task, this task can also be divided into intra- and inter-document sub-tasks. For the intra-document sub-task, the instances are all sampled from one single document. For the inter-document sub-task, the instances are sampled from different documents. 

\smallskip
\noindent
\textbf{Relational Fact Alignment.} To grasp useful information from the long documents and denoise irrelevant content, we design the relation-level task, which requires the representations of the same entity pairs in different documents to be similar.
Formally, assume $d_A$ and $d_B$ are two documents from the training set, which share several relational facts. Let $\{r_A^i\}_{i=1}^{k_s}$ denote the relational instances in $d_A$, and $r_B^q$ denote the representation of the relational instance in $d_B$ whose relational fact is contained in $d_A$. Then the model is required to find the relational instance from $\{r_A^i\}_{i=1}^{k_s}$, which shares the same relational fact with $r_B^q$. First, we compute the similarity score of two relational instances:
\begin{equation}
    s_s(r_A^i, r_B^q) = \mathbf{w}_s|\mathbf{r}_A^i - \mathbf{r}_{B}^q| + b_s.
\end{equation}
$\mathbf{w}_s$ and $b_s$ are weights and bias. Then similarity scores are fed into a softmax over instances in $d_A$.

Finally, the overall pre-training loss $\mathcal{L}$ is the sum of all cross-entropy losses in three tasks. 
\begin{equation}
    \mathcal{L} = \mathcal{L}_M + \mathcal{L}_S + \mathcal{L}_N.
    \label{loss_function}
\end{equation}

Note that the loss can be easily minimized by an entity linking system without any relational knowledge. To avoid this problem, we replace all the mentions of an entity in a document by a special blank symbol \texttt{[BLANK]} with probability $\alpha$ following \citet{soares2019matching}. In such a case, the model can only learn representations from the context. As a result, minimizing the loss $\mathcal{L}$ requires the model to do more than just memorizing named entities.

\subsection{Pre-denoising Module}

As stated before, the document-level DS will generate more noise. To alleviate the issue, we propose to screen out entity pairs with low relational probability from all documents with a rank model.
We train the rank model with the Relation Detection task on the human-annotated training set. Then, the rank model is able to give high scores to positive instances and low scores to NA instances. During the pre-denoising process, we compute positive scores for all entity pairs as stated in Eq.~\ref{score:positive}. Next, for each document, we rank all its entity pairs according to their positive scores, and keep top $k_d$ entity pairs for pre-training, fine-tuning and evaluation. The framework of the pre-denoising module is the same as the model used for pre-training. Please refer to the previous section for details.
With the pre-denoising module, the wrong labeling problem in DS corpus and the label imbalance problem (i.e., most entity pairs belong to NA instances) in the human-annotated corpus can be alleviated.


\section{Experiments}
\subsection{Dataset and Evaluation Metrics}
We evaluate our model on DocRED~\cite{yao2019docred}, which is the largest human-annotated DocRE dataset.
DocRED contains $5,053$ human-annotated documents, with $56,354$ relational facts and $63,427$ relation instances. Besides, DocRED also provides large-scale distantly supervised data, which contains $101,873$ documents, with $881,298$ relational facts and $1,508,320$ relation instances labeled automatically by distant supervision~\cite{mintz2009distant}. Please refer to \citet{yao2019docred} for the details about dataset construction.

In experiments, we use the document-level DS data to pre-train our model and then fine-tune and evaluate the model on the human-annotated data. Following~\citet{yao2019docred}, we use $F_1$ and $IgnF_1$ as evaluation metrics, where $IgnF_1$ denote the $F_1$ scores excluding relational facts in both training and dev/test sets. Please refer to the appendix for details about DocRED and experimental settings.

\begin{table}
\centering
\small
\begin{tabular}{lcccc}
\toprule
             & \multicolumn{2}{c}{Dev} & \multicolumn{2}{c}{Test} \\
Model        & $F_1$      & $IgnF_1$   & $F_1$      & $IgnF_1$      \\ \midrule
CNN*          & 43.45      & 41.58      & 42.26      & 40.33       \\
LSTM*         & 48.44      & 50.68      & 47.71      & 50.07       \\
BiLSTM*       & 50.94      & 48.87      & 51.06      & 48.78       \\
ContextAware* & 51.09      & 48.94      & 50.70      & 48.40       \\
BERT          & 55.67      & 53.32      & 56.17      & 53.66          \\
BERT-TS $\clubsuit$ & 54.42      & --         & 53.92      & --          \\
HIN-BERT $\spadesuit$      & 56.31      & 54.29      & 55.60      & 53.70       \\
\midrule
DS-BiLSTM*       & 51.72      & 41.44      & 49.80      & 39.15       \\
DS-ContextAware* & 51.39      & 40.47      & 50.12      & 39.16       \\
\midrule
BERT+D        & 57.42      & 55.88      & 57.20      & 55.53       \\
BERT+D+P    & \textbf{58.65}  & \textbf{57.00}  & \textbf{58.43}  & \textbf{56.68} \\
\bottomrule
\end{tabular}
\caption{Main results on DocRED. Results with *, $\clubsuit$ and $\spadesuit$ are from \citet{yao2019docred}, \citet{wang2019fine} and \citet{tang2020hin} respectively.}
\label{main_result}
\end{table}

\subsection{Baseline}
We compare our model with the following baselines.
(1) CNN/LSTM/BiLSTM~\cite{yao2019docred}: these models capture relational semantics via various encoder.
(2) ContextAware~\cite{sorokin2017context}: it considers the relations' interactions with attention to jointly learn all entity pairs in the contexts. 
(3) BERT~\cite{devlin2019bert}: this baseline is implemented as described in Sec.~\ref{sec:encoder}. 
(4) BERT-TS~\cite{wang2019fine}: 
it predicts whether two entities have relations in the first step and then predicts the specific relation in the second step.
(5) HIN-BERT~\cite{tang2020hin}: it applies a hierarchical inference network to aggregate information from multiple granularity.
(6) DS-BiLSTM/ContextAware~\cite{yao2019docred}: corresponding models trained on DS data.

\subsection{Implementation Details}
We pre-train our model based on $\text{BERT}_\text{BASE}$. All the hyper-parameters are selected with manually tuning. The learning rate is set to $3 \times 10^{-5}$ for pre-training and $10^{-5}$ for fine-tuning. The size of relational representations is $256$, which is selected from $\{64, 128, 256, 512\}$. The batch size for pre-training is set to $16$ and $4$ for fine-tuning. We keep $2N_{\text{ent}}$ entity pairs after pre-denoising for each document during fine-tuning, where $N_{\text{ent}}$ is the number of entities mentioned in the document. And we keep $20$ entity pairs for each document during pre-training. We train our model with GeForce RTX 2080 Ti. All the special tokens including entity markers and the special blank symbol are implemented with unused tokens in the $\text{BERT}_\text{BASE}$ vocabulary.  

\subsection{Main Result}
The main results are shown in Tab.~\ref{main_result}. Specifically, \textbf{D} refers to the pre-denoising module and \textbf{P} indicates pre-training tasks. From the results, we can observe that:
(1) Our model outperforms all baselines by a significant margin. 
It is due to the effectiveness of the pre-denoising mechanism and three pre-training tasks. (2) Our model without pre-training (BERT+D) can also outperform all the baseline models, which indicates that our pre-denoising module can deal with the amounts of NA instances.
(3) The noise in distantly supervised data harms the performance of RE systems. Our model can filter out noise and capture information from the large-scale distantly supervised data, thus achieving a performance improvement.
(4) Pre-training without pre-denoising (BERT+P) cannot converge due to amounts of data labeled incorrect.



\begin{table}
\centering
\small
\begin{tabular}{lcccc}
\toprule
             & \multicolumn{2}{c}{Dev} & \multicolumn{2}{c}{Test} \\
Model        & $F_1$      & $IgnF_1$   & $F_1$      & $IgnF_1$      \\ \midrule
our model   & 58.65  & \textbf{57.00}  & \textbf{58.43}  & \textbf{56.68} \\
\midrule
\quad w/o MM    & 58.39      & 56.76      & 57.60      & 55.81            \\
\quad w/o RD    & 57.19      & 55.61      & 56.71      & 54.94            \\
\quad w/o RA    & 58.48      & 56.73      & 58.13      & 56.30            \\
\midrule
\quad w/o Inter & \textbf{58.68}  & 56.96      & 57.72      & 55.87             \\
\quad w/o Intra & 57.78      & 56.18      & 57.62      & 55.89              \\
\bottomrule
\end{tabular}
\caption{Results of ablation study on DocRED.}
\label{ablation_study}
\end{table}

\subsection{Ablation Study}
To explore the contribution of different pre-training tasks, we show the results of the ablation study in Tab.~\ref{ablation_study}. Specifically, we show the scores with different pre-training tasks turned off one at a time. \textbf{MM}, \textbf{RD}, \textbf{RA} indicate three pre-training tasks: Mentions/Entities Matching, Relation Detection, and Relational Facts Alignment. We observe that all three pre-training tasks contribute to the main model, as the performance deteriorates with any of the tasks missing. Note that the removal of the RD pre-training task leads to a large drop in both $F_1$ and $IgnF_1$ scores, even lower than those of our model without pre-training (BERT+D). This is because without RD, the model is unable to identify positive instances, which is quite important in document-level RE and then the label imbalance problem makes the scores drop.

Moreover, we conduct another ablation study to explore the effectiveness of intra- and inter-document subtasks. The results are shown in Tab.~\ref{ablation_study}, where \textbf{w/o Intra} and \textbf{w/o Inter} refer to pre-training without intra- and inter-document sub-tasks. We find that both intra-document and inter-document sub-tasks contribute to the main model in general.

\section{Conclusion}
In this work, we propose to denoise distantly supervised data in DocRE by multiple pre-training tasks. Experiment results verify the effectiveness of our model. In the future, we will explore how to improve the efficiency of our pre-training.

\section*{Acknowledgement}
This work is supported by the National Key Research and Development Program of China (No. 2018YFB1004503), the National Natural Science Foundation of China (NSFC No. 61532010) and Beijing Academy of Artificial Intelligence (BAAI).

\bibliography{anthology,emnlp2020}

\begin{thebibliography}{20}
\expandafter\ifx\csname natexlab\endcsname\relax\def\natexlab#1{#1}\fi

\bibitem[{Christopoulou et~al.(2019)Christopoulou, Miwa, and
  Ananiadou}]{christopoulou2019connecting}
Fenia Christopoulou, Makoto Miwa, and Sophia Ananiadou. 2019.
\newblock \href {https://www.aclweb.org/anthology/D19-1498.pdf} {Connecting the
  dots: Document-level neural relation extraction with edge-oriented graphs}.
\newblock In \emph{Proceedings of EMNLP}.

\bibitem[{Devlin et~al.(2019)Devlin, Chang, Lee, and
  Toutanova}]{devlin2019bert}
Jacob Devlin, Ming-Wei Chang, Kenton Lee, and Kristina Toutanova. 2019.
\newblock \href {https://www.aclweb.org/anthology/N19-1423.pdf} {Bert:
  Pre-training of deep bidirectional transformers for language understanding}.
\newblock In \emph{Proceedings of NAACL}.

\bibitem[{Feng et~al.(2018)Feng, Huang, Zhao, Yang, and
  Zhu}]{DBLP:conf/aaai/FengHZYZ18}
Jun Feng, Minlie Huang, Li~Zhao, Yang Yang, and Xiaoyan Zhu. 2018.
\newblock \href
  {https://www.aaai.org/ocs/index.php/AAAI/AAAI18/paper/viewPDFInterstitial/17151/16140}
  {Reinforcement learning for relation classification from noisy data}.
\newblock In \emph{Proceedings of AAAI}, pages 5779--5786.

\bibitem[{Gu et~al.(2017)Gu, Sun, Qian, and Zhou}]{gu2017chemical}
Jinghang Gu, Fuqing Sun, Longhua Qian, and Guodong Zhou. 2017.
\newblock \href
  {https://academic.oup.com/database/article/doi/10.1093/database/bax024/3098440}
  {Chemical-induced disease relation extraction via convolutional neural
  network}.
\newblock \emph{Database}.

\bibitem[{Lin et~al.(2016)Lin, Shen, Liu, Luan, and Sun}]{lin2016neural}
Yankai Lin, Shiqi Shen, Zhiyuan Liu, Huanbo Luan, and Maosong Sun. 2016.
\newblock \href {https://www.aclweb.org/anthology/P16-1200.pdf} {Neural
  relation extraction with selective attention over instances}.
\newblock In \emph{Proceedings of ACL}.

\bibitem[{Min et~al.(2013)Min, Grishman, Wan, Wang, and
  Gondek}]{min2013distant}
Bonan Min, Ralph Grishman, Li~Wan, Chang Wang, and David Gondek. 2013.
\newblock \href {https://www.aclweb.org/anthology/N13-1095.pdf} {Distant
  supervision for relation extraction with an incomplete knowledge base}.
\newblock In \emph{Proceedings of the NAACL-HLT}, pages 777--782.

\bibitem[{Mintz et~al.(2009)Mintz, Bills, Snow, and
  Jurafsky}]{mintz2009distant}
Mike Mintz, Steven Bills, Rion Snow, and Dan Jurafsky. 2009.
\newblock \href {https://dl.acm.org/ft_gateway.cfm?ftid=878759&id=1690287}
  {Distant supervision for relation extraction without labeled data}.
\newblock In \emph{Proceedings of ACL}.

\bibitem[{Peng et~al.(2017)Peng, Poon, Quirk, Toutanova, and
  Yih}]{peng2017cross}
Nanyun Peng, Hoifung Poon, Chris Quirk, Kristina Toutanova, and Wen-tau Yih.
  2017.
\newblock \href {https://www.mitpressjournals.org/doi/pdf/10.1162/tacl_a_00049}
  {Cross-sentence n-ary relation extraction with graph lstms}.
\newblock \emph{TACL}.

\bibitem[{Qin et~al.(2018)Qin, Xu, and Wang}]{qin2018dsgan}
Pengda Qin, Weiran Xu, and William~Yang Wang. 2018.
\newblock \href {https://www.aclweb.org/anthology/P18-1046.pdf} {Dsgan:
  generative adversarial training for distant supervision relation extraction}.
\newblock In \emph{Proceedings of ACL}.

\bibitem[{Quirk and Poon(2017)}]{quirk2017distant}
Chris Quirk and Hoifung Poon. 2017.
\newblock \href {https://www.aclweb.org/anthology/E17-1110.pdf} {Distant
  supervision for relation extraction beyond the sentence boundary}.

\bibitem[{Sahu et~al.(2019)Sahu, Christopoulou, Miwa, and
  Ananiadou}]{sahu2019inter}
Sunil~Kumar Sahu, Fenia Christopoulou, Makoto Miwa, and Sophia Ananiadou. 2019.
\newblock \href {https://www.aclweb.org/anthology/P19-1423.pdf} {Inter-sentence
  relation extraction with document-level graph convolutional neural network}.
\newblock In \emph{Proceedings of ACL}.

\bibitem[{Soares et~al.(2019)Soares, FitzGerald, Ling, and
  Kwiatkowski}]{soares2019matching}
Livio~Baldini Soares, Nicholas FitzGerald, Jeffrey Ling, and Tom Kwiatkowski.
  2019.
\newblock \href {https://www.aclweb.org/anthology/P19-1279.pdf} {Matching the
  blanks: Distributional similarity for relation learning}.
\newblock In \emph{Proceedings of ACL}, pages 2895--2905.

\bibitem[{Sorokin and Gurevych(2017)}]{sorokin2017context}
Daniil Sorokin and Iryna Gurevych. 2017.
\newblock \href {https://www.aclweb.org/anthology/D17-1188.pdf} {Context-aware
  representations for knowledge base relation extraction}.
\newblock In \emph{Proceedings of EMNLP}.

\bibitem[{Takanobu et~al.(2019)Takanobu, Zhang, Liu, and
  Huang}]{takanobu2019hierarchical}
Ryuichi Takanobu, Tianyang Zhang, Jiexi Liu, and Minlie Huang. 2019.
\newblock \href
  {https://www.aaai.org/ojs/index.php/AAAI/article/download/4688/4566} {A
  hierarchical framework for relation extraction with reinforcement learning}.
\newblock In \emph{Proceedings of AAAI}.

\bibitem[{Tang et~al.(2020)Tang, Cao, Zhang, Cao, Fang, Wang, and
  Yin}]{tang2020hin}
Hengzhu Tang, Yanan Cao, Zhenyu Zhang, Jiangxia Cao, Fang Fang, Shi Wang, and
  Pengfei Yin. 2020.
\newblock \href
  {https://link.springer.com/chapter/10.1007/978-3-030-47426-3_16} {Hin:
  Hierarchical inference network for document-level relation extraction}.
\newblock In \emph{Proceedings of PAKDD}, pages 197--209.

\bibitem[{Wang et~al.(2019)Wang, Focke, Sylvester, Mishra, and
  Wang}]{wang2019fine}
Hong Wang, Christfried Focke, Rob Sylvester, Nilesh Mishra, and William Wang.
  2019.
\newblock \href {https://arxiv.org/pdf/1909.11898} {Fine-tune bert for docred
  with two-step process}.
\newblock \emph{arXiv preprint arXiv:1909.11898}.

\bibitem[{Xu et~al.(2016)Xu, Wu, Zhang, Wang, Lee, and Xu}]{xu2016cd}
Jun Xu, Yonghui Wu, Yaoyun Zhang, Jingqi Wang, Hee-Jin Lee, and Hua Xu. 2016.
\newblock \href
  {https://academic.oup.com/database/article/doi/10.1093/database/baw036/2630291}
  {Cd-rest: a system for extracting chemical-induced disease relation in
  literature}.
\newblock \emph{Database}, 2016.

\bibitem[{Yao et~al.(2019)Yao, Ye, Li, Han, Lin, Liu, Liu, Huang, Zhou, and
  Sun}]{yao2019docred}
Yuan Yao, Deming Ye, Peng Li, Xu~Han, Yankai Lin, Zhenghao Liu, Zhiyuan Liu,
  Lixin Huang, Jie Zhou, and Maosong Sun. 2019.
\newblock \href {https://www.aclweb.org/anthology/P19-1074.pdf} {Docred: A
  large-scale document-level relation extraction dataset}.
\newblock In \emph{Proceedings of ACL}.

\bibitem[{Zeng et~al.(2014)Zeng, Liu, Lai, Zhou, Zhao
  et~al.}]{zeng2014relation}
Daojian Zeng, Kang Liu, Siwei Lai, Guangyou Zhou, Jun Zhao, et~al. 2014.
\newblock \href {https://www.aclweb.org/anthology/C14-1220.pdf} {Relation
  classification via convolutional deep neural network}.
\newblock In \emph{Proceedings of COLING}.

\bibitem[{Zhou et~al.(2018)Zhou, Xu, Qi, Bao, Chen, and Xu}]{zhou2018distant}
Peng Zhou, Jiaming Xu, Zhenyu Qi, Hongyun Bao, Zhineng Chen, and Bo~Xu. 2018.
\newblock \href
  {https://www.sciencedirect.com/science/article/abs/pii/S0893608018302429}
  {Distant supervision for relation extraction with hierarchical selective
  attention}.
\newblock \emph{Neural Networks}.

\end{thebibliography}
\bibliographystyle{acl_natbib}

\end{document}